\documentclass{article}
\usepackage{spconf,amsmath,graphicx}
\usepackage{amsfonts}
\usepackage{amsmath}
\usepackage{adjustbox}
\usepackage{algorithmicx}
\usepackage{algpseudocode}
\usepackage[ruled]{algorithm}
\usepackage{graphicx}
\usepackage{caption}
\usepackage{graphicx,times,amsmath} 
\usepackage{caption}
\usepackage{subcaption}
\usepackage[rightcaption]{sidecap}
\usepackage{caption}
\usepackage{multirow}
\usepackage{hhline}
\usepackage{amsmath}
\usepackage{epstopdf}
\usepackage{tabularx}
\usepackage{tikz}
\usetikzlibrary{matrix}
\usepackage{mathtools, nccmath}
\usepackage{pst-node}

\graphicspath{ {images/} }

\usepackage{fancyhdr}

\usepackage{alphalph}

\title{A Framework For Pruning Deep Neural Networks Using\\ Energy-based Models}

\name{Hojjat Salehinejad, Member, IEEE, and Shahrokh Valaee, Fellow, IEEE}
\address{Department of Electrical \& Computer Engineering, University of Toronto, Toronto, Canada \\
\textit{hojjat.salehinejad@mail.utoronto.ca, valaee@ece.utoronto.ca}}

\usepackage{fancyhdr}
\usepackage{fancyhdr}
\fancypagestyle{pageStyleOne}{%
\normalfont\footnotesize
    \fancyhf{}
\fancyhead[L]{\fontsize{8}{8} \selectfont This paper is accepted for presentation at IEEE International Conference on Acoustics, Speech and Signal Processing (IEEE ICASSP), 2021.}}

\begin{document}
\newcommand*{\img}{%
  \includegraphics[
    width=\linewidth,
    height=20pt,
    keepaspectratio=false,
  ]{example-image-a}%
}

\maketitle
\thispagestyle{pageStyleOne}

\begin{abstract}
A typical deep neural network (DNN) has a large number of trainable parameters. Choosing a network with proper capacity is challenging and generally a larger network with excessive capacity is trained. Pruning is an established approach to reducing the number of parameters in a DNN. In this paper, we propose a framework for pruning DNNs based on a population-based global optimization method. This framework can use any pruning objective function. As a case study, we propose a simple but efficient objective function based on the concept of energy-based models. Our experiments on ResNets, AlexNet, and SqueezeNet for the CIFAR-10 and CIFAR-100 datasets show a pruning rate of more than $50\%$ of the trainable parameters with approximately $<5\%$ and $<1\%$ drop of Top-1 and Top-5 classification accuracy, respectively.
\end{abstract}
\section{Introduction}
\label{sec:intro}
Pruning a deep neural network (DNN) is one of the major methods for removing redundant trainable parameters and compressing the network. This approach permanently removes a subset of trainable parameters. In general, pruning algorithms have three stages which are training, pruning, and fine-tuning~\cite{liu2018rethinking}. One pruning approach is to utilize second derivative information to minimize a cost function that reduces network complexity by removing excess number of trainable parameters and further fine-tuning~\cite{lecun1990optimal}. One of the major approaches is \textit{Deep Compression}, which has three main steps that are pruning, quantization, and Huffman coding~\cite{han2015deep}. It prunes all connections with weights below a given threshold and then retrains the sparsified network.

\begin{figure}[t]
\centering
\captionsetup{font=footnotesize}
\includegraphics[width=0.4\textwidth]{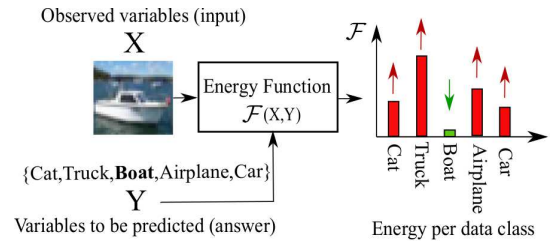}  
\caption{Energy-based model (EBM),~\cite{lecun2006tutorial}.}
\label{fig:energy_model_diagram}
\vspace{-6mm}
\end{figure}

 Generally, probabilistic models can be considered as a special type of energy-based models (EBMs). An EBM assigns a \textit{scalar energy loss} as a measure of compatibility to a configuration of parameters in neural networks as demonstrated in Figure~\ref{fig:energy_model_diagram}. This approach avoids computing the normalization term, which can be interpreted as an alternative to probabilistic estimation~\cite{lecun2006tutorial}.
 Calculating the exact probability needs computing the partition function over all the data classes. However, for large number of data classes, such as in language models with more than $100,000$ classes, this causes a bottleneck~\cite{barber2016dealing}. Some methods such as annealed importance sampling~\cite{neal2001annealed} have been proposed to deal with this problem which is out of the scope of this paper.

Previously, we have proposed an Ising energy model for dropout and pruning of multilayer perceptron (MLP) networks~\cite{salehinejad2019ising, salehinejad2019isingGlobal}. In this paper, a pruning framework based on a population-based stochastic global optimization method is proposed which is integrated into the typical training procedure of a DNN. This scheme is inspired from the concept of dropout~\cite{labach2019survey} and biological pruning of neurons in brain. This framework can handle different pruning objective function with multiple constrains. We also propose an energy-based pruning objective function based on the concept of EBM in DNN, which allocates a scalar energy value to each state vector in the population of state vectors, called \textit{EPruning}. Each state vector is in fact a representation of a sub-network from the original DNN. Pruning is defined as searching for a binary state vector that prunes the network while minimizes the energy loss for a set of inputs and corresponding outputs in each iteration. Hence, the search for weights is conducted using a probabilistic model while the pruning state vector is fixed and the search for pruning state is conducted using an EBM while the weights are fixed in each iteration. The candidate states help to find a subset of the neural network and capture its energy function that associates low energies to correct values of the remaining variables, and higher energies to incorrect values\footnote{The codes and more details of experiments setup is available at: \textit{https://github.com/sparsifai/epruning}}.

\section{Proposed Pruning Framework}

\subsection{Energy Model} 
\begin{figure}
\centering
\captionsetup{font=footnotesize}
\includegraphics[width=0.35\textwidth]{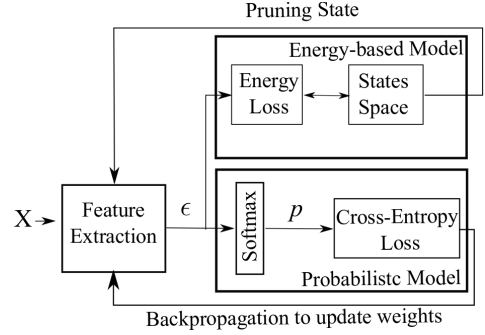}  
\caption{Switching between the energy-based model (EBM) and the probabilistic model. The EBM searches for the pruning state and the probabilistic models searches for the weights. Both models are aware of target class $Y$ during training. In inference, the best pruning state is applied and the EBM is removed.}
\label{fig:energy_model_block}
\vspace{-6mm}
\end{figure}

A DNN can be modeled as a parametric function to map the input image $X\in\mathcal{X}$ to $C$ real-valued numbers $\epsilon=\{\varepsilon_{1},\varepsilon_{2},...,\varepsilon_{C}\}$ (a.k.a. logits). The output is then passed to a classifier, such as \textit{Softmax} function to parameterize a categorical distribution in form of a probability distribution over the data classes $\mathcal{Y}=\{y_{1},...,y_{C}\}$~\cite{grathwohl2019your}, defined as $\{p(y_{1}),...,p(y_{C})\}$ where for simplicity we define $p_{c}=p(y_{c})\:\forall \: c\in\{1,...,C\}$, as illustrated in Figure~\ref{fig:energy_model_block}.
The loss is then calculated based on cross-entropy with respect to the correct answer $Y$. 
Gibbs distribution is a very general family of probability distributions defined as 
\begin{equation}
p(Y|X)=\frac{e^{-\beta \mathcal{F}(Y,X)}}{Z(\beta)},
\label{eq:gibbs}
\end{equation}
where $Z(\beta)=\sum\limits_{y_{c}\in\mathcal{Y}}e^{-\beta \mathcal{F}(y_{c},X)}$ is the partition function, ${\beta>0}$ is the inverse temperature parameter~\cite{lecun2006tutorial}, and $\mathcal{F}(\cdot)$ is the \textit{Hamiltonian} or the \textit{energy function}. 
Softmax function is a special case of the Gibbs distribution. We can achieve the energy function corresponding to using a Softmax layer by setting $\beta=1$~\cite{murphy2012machine} in~(\ref{eq:gibbs}) and getting the Hamiltonian 
\begin{equation}
\mathcal{F}(Y,X) = -\epsilon.   
\end{equation}
We define the following energy loss function to measure the quality of energy function for $(X,Y)$ with target output $y_{c}$ as
\begin{equation}
\begin{split}
\mathcal{E}&=\mathcal{L}(Y,\mathcal{F})\\
&=\mathcal{F}(y_{c},X)-
min\{\mathcal{F}(y_{c'},X): \; y_{c'}\in\mathcal{Y},c'\neq c\},
\end{split}
\label{eq:energy_loss_function}
\end{equation}
where it can be extended for a batch of data.
The \textit{energy loss} function assigns a low loss value to a pruning state vector which has the lowest energy with respect to the target data class $c$ and higher energy with respect to the other data classes and vice versa~\cite{lecun2006tutorial}. 

\subsection{Training and Optimization}
We are interested in pruning weight kernels and hidden units, including their bias terms, referred to as a \textit{unit} hereafter for convenience. 
We define a set of $S$ binary candidate pruning state vectors as the population $\mathbf{S}^{S\times D}$. Each vector $\mathbf{s}_{i}\in\mathbf{S}$ has a length of $D$ which represents a sub-network. The energy function value for each state vector is $\mathcal{F}_{i}\in\{\mathcal{F}_{1},...,\mathcal{F}_{S}\}$. If $s_{i,d}=0$ unit $d$ is dropped and if $s_{i,d}=1$ it is active. 

Algorithm~\ref{alg:EPruning} shows different steps of the proposed framework. 
At the beginning of training ($t=0$), we initialize the candidate pruning states $\mathbf{S}^{(0)}\in \mathbb{Z}_{2}^{S\times D}$, where $s_{i,d}^{(0)}\sim Bernoulli(P=0.5)$ for $i\in\{1,...,S\}$ and $d\in\{1,...,D\}$.

For each candidate state $\mathbf{s}_{i}^{(t)}\in\mathbf{S}^{(t)}$ in iteration $t$, the energy loss value is calculated using~(\ref{eq:energy_loss_function}) as $\mathcal{E}^{(t)}_{i}$.
Searching for the pruning state which can minimize the energy loss value is an NP-hard combinatorial problem. Various methods such as 
MCMC~\cite{dabiri2020replica} and simulated annealing (SA) can be used to search for low energy states. We propose using a binary version of differential evolution (BDE)~\cite{price2013differential} to minimize the energy loss function. This method has the advantage of searching the optimization landscape in parallel and sharing the search experience among candidate states. The other advantage of this approach is flexibility of designing the energy function with constraints.

The optimization step has three phases which are mutation, crossover, and selection. Given the population of states $\mathbf{S}^{(t-1)}$, a mutation vector is defined for each candidate state ${\mathbf{s}_{i}^{(t-1)}\in\mathbf{S}^{(t-1)}}$ as 
\begin{equation}
v_{i,d}=\begin{cases}
               1-s_{i_{1},d}^{(t-1)}, \:\:\:\:$if$\:\:\:s_{i_{2},d}^{(t-1)}\neq s_{i_{3},d} ^{(t-1)}\;$\&$\; r_{d}<F\\
               s_{i_{1},d}^{(t-1)}, \:\:\:\:\:\:\:\:\:\:\:$ otherwise $
            \end{cases},
\label{eq:mutation}
\end{equation}
for all $d\in\{1,..,D\}$, where $i_{1},i_{2},i_{3}\in \{1,...,S\}$ are mutually different, $F$ is the mutation factor~\cite{salehinejad2017micro}, and $r_{d}\in[0,1]$ is a random number. The next step is to crossover the mutation vectors to generate new candidate state vectors as
\begin{equation}
\tilde{s}^{(t)}_{i,d}=\begin{cases}
               v_{i,d} \:\:\:\:\:\:\:\:\:\:\:\:$if$\:\:\: r'_{d}\in[0,1] \leq C_{r}\\
               s_{i,d}^{(t-1)} \:\:\:\:\:\:\:\:\:\:\:$ otherwise $
            \end{cases},
\label{eq:crossover}
\end{equation}
where $C_{r}$ is the crossover coefficient~\cite{salehinejad2017micro}. The parameters $C_{r}$ and $F$ control exploration and exploitation of the population on the optimization landscape. Each generated state $ \tilde{\mathbf{s}}_{i}^{(t)}$ is then compared with its corresponding parent with respect to its energy loss value $\tilde{\mathcal{E}}^{(t)}_{i}$ as
\begin{equation}
\mathbf{s}_{i}^{(t)}=\begin{cases}
               \tilde{\mathbf{s}}_{i}^{(t)} \:\:\:\:\:\:\:\:\:\:$if$\:\:\:    \tilde{\mathcal{E}}^{(t)}_{i}\leq \mathcal{E}^{(t-1)}_{i} \\
              \mathbf{s}_{i}^{(t-1)} \:\:\:\:\:$ otherwise $
            \end{cases}\:\forall\:i\in\{1,...,S\}.
\label{eq:selection}
\end{equation}
The state with minimum energy loss $\mathcal{E}_{b}^{(t)}=min\{\mathcal{E}_{1}^{(t)},...,\mathcal{E}_{S}^{(t)}\}$ is selected as the best state $\mathbf{s}_{b}$, which represents the sub-network for next training batch. This optimization strategy is simple and feasible to implement in parallel for a large $S$.  

The population-based optimization methods suffer from premature convergence and stagnation problems. The former generally occurs when the population (candidate state vectors) has converged to local optima, has lost its diversity, or has no improvement in finding better solutions. The latter happens mainly when the population stays diverse during training~\cite{lampinen2000stagnation}.
After a number of iterations, depending on the capacity of the neural network and the complexity of the dataset, all the states in $\mathbf{S}^{(t)}$ may converge to a state $\mathbf{s}_{b}\in\mathbf{S}^{(t)}$. We call this the \textit{early state convergence} phase and define it as 
\begin{equation} 
\Delta\mathbf{s} =
\mathcal{E}_{b}^{(t)} -
\frac{1}{S}\sum\limits_{j=1}^{S}\mathcal{E}_{j}^{(t)},
\label{eq:stateconvergence}
\end{equation}
where $\mathcal{E}_{b}^{(t)}$ is the energy loss of $\mathbf{s}_{b}$.
Therefore, if $\Delta\mathbf{s}=0$ we can call for an {early state convergence} and continue training by fine-tuning the sub-network identified by the state vector $\mathbf{s}_{b}$. In addition, a stagnation threshold $\Delta\mathbf{s}_{T}$ is implemented where if $\Delta\mathbf{s}\neq 0$ after $\Delta\mathbf{s}_{T}$ number of training epochs, it stops the energy loss optimizer and starts fine-tuning the selected sub-network. The convergence to the best state $\mathbf{s}_{b}$ splits the training procedure into two phases, where the first phase acts similar to dropout and the second phase fine-tunes the pruned network, defined by $\mathbf{s}_{b}$.

\begin{algorithm}[t]
\small
\begin{algorithmic} 
\State Set $t$ = 0 // Optimization counter
\State Initiate the neural network with trainable weights $\Theta$
\State Set $\mathbf{S}^{(0)}\sim Bernoulli(P=0.5)$ // States initialization
\State Set $\Delta\mathbf{s}\neq 0$ \& $\Delta\mathbf{s}_{T}$
\For{ $ \mathit{i_{epoch}} = 1 \rightarrow \mathit{N_{epoch}}$} // epoch counter
\For{ $ \mathit{i_{batch}} = 1 \rightarrow \mathit{N_{batch}}$} // batch counter
\State $t = t+1$
\If{$\Delta\mathbf{s}\neq 0$ or $i_{epoch}\leq \Delta\mathbf{s}_{T}$}
\If{$i_{epoch}=1 \; \& \; i_{batch}=1$}
\State Compute energy loss of $\mathbf{S}^{(0)}$ as $\mathcal{E}^{(0)}$ using~(\ref{eq:energy_loss_function})
\EndIf

\For {$i=1\rightarrow S$} // States counter
\State Generate mutually different $i_{1},i_{2},i_{3}\in \{1,...,S\}$
\For {$d=1\rightarrow D$} // State dimension counter
\State Generate a random number $r_{d}\in[0,1]$
\State Compute mutation vector $v_{i,d}$ using (\ref{eq:mutation})
\State Compute candidate state $\tilde{s}^{(t)}$ using (\ref{eq:crossover})
\EndFor
\EndFor
\State Compute energy loss of $\tilde{\mathbf{S}}^{(t)}$ as $\tilde{\mathcal{E}}^{(t)}$ using~(\ref{eq:energy_loss_function})
\State Select $\mathbf{S}^{(t)}$ and corresponding energy $\mathcal{E}^{(t)}$ using (\ref{eq:selection})
\State Select the state with the lowest energy from $\mathbf{S}^{(t)}$ as $\mathbf{s}^{(t)}_{b}$
\Else
\State $\mathbf{s}^{(t)}_{b}=\mathbf{s}^{(t-1)}_{b}$
\EndIf
\State Temporarily drop weights of the network based on the best state $\mathbf{s}^{(t)}_{b}$
\State Compute loss of the sparsified network
\State Perform backpropagation to update $\Theta$
\EndFor
\State Update $\Delta\mathbf{s}$ for early state convergence using (\ref{eq:stateconvergence})
\EndFor
\end{algorithmic}
\small
  \caption{EPruning}
  \label{alg:EPruning}
\end{algorithm}

\begin{figure*}[!t]
\centering
\captionsetup{font=footnotesize}
\begin{subfigure}[t]{0.49\textwidth}
\captionsetup{font=footnotesize}
\centering
\includegraphics[width=1\textwidth]{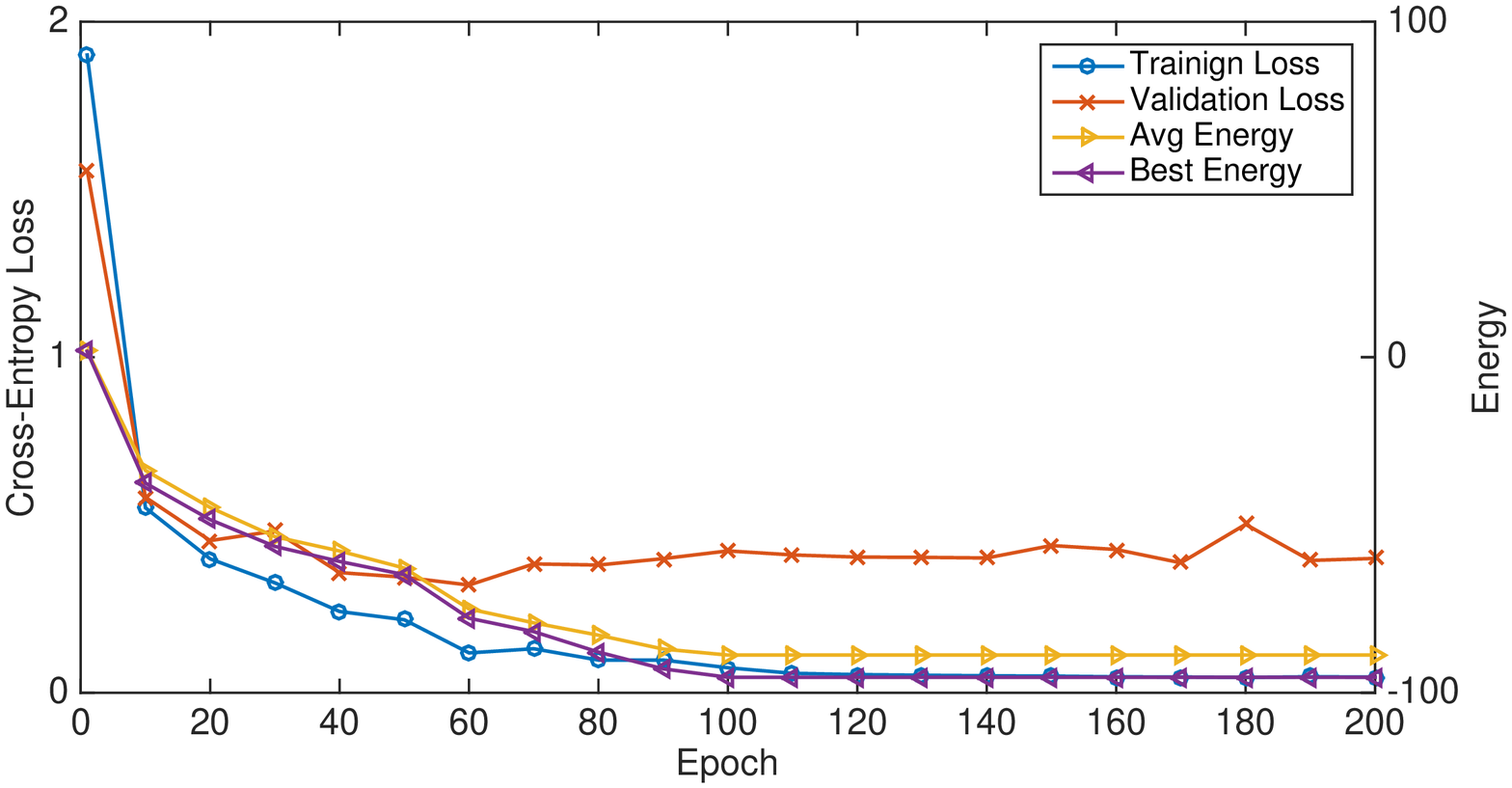}  
\caption{CIFAR-10}
\end{subfigure}%
~
\begin{subfigure}[t]{0.49\textwidth}
\captionsetup{font=footnotesize}
\centering
\includegraphics[width=1\textwidth]{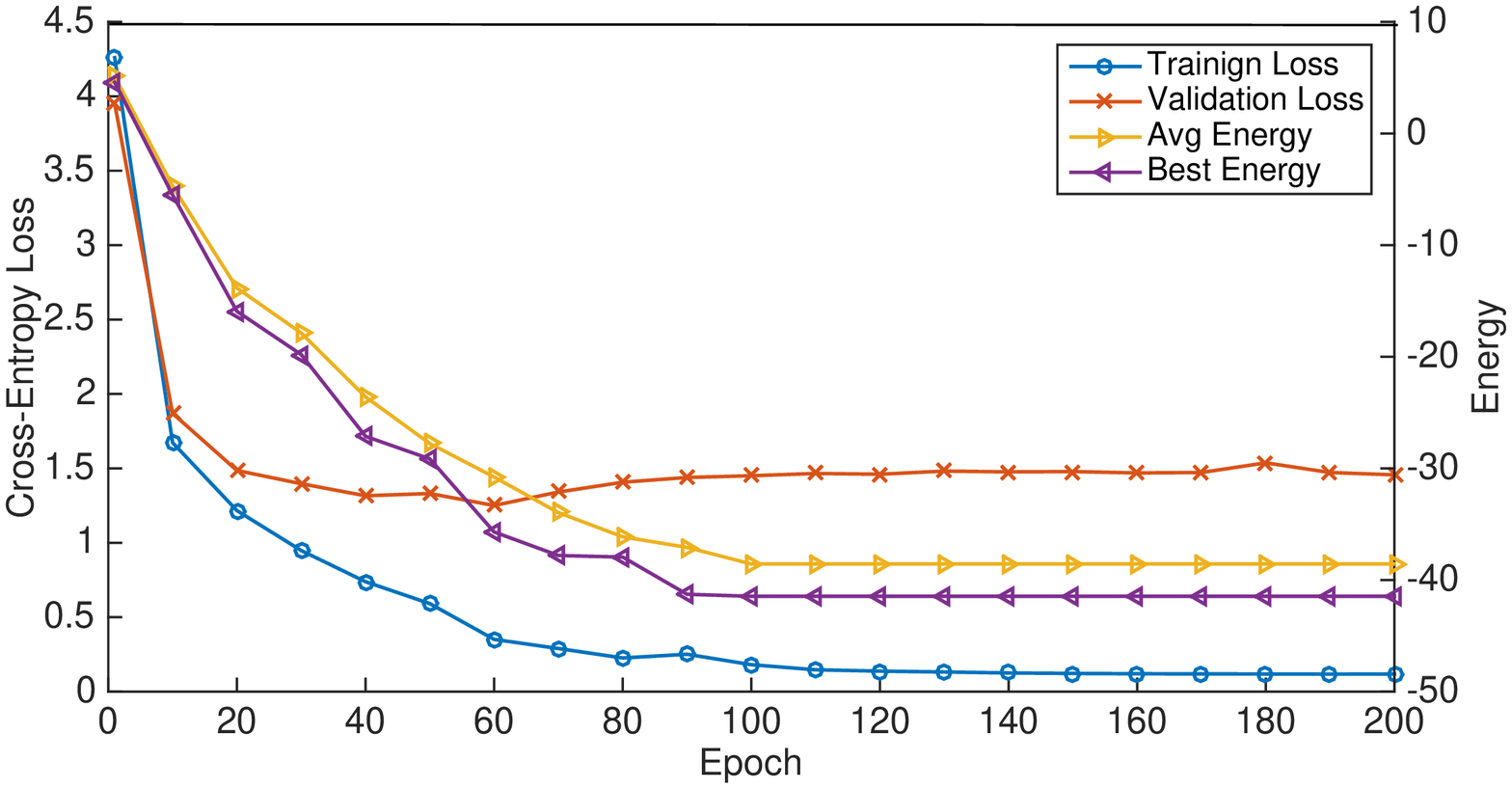}
\caption{CIFAR-100}
\end{subfigure}   
\caption{Cross-Entropy loss and energy of \textit{EPruning} over 200 training epochs of \textbf{ResNet-50} for CIFAR-10 and CIFAR-100 datasets with $S=8$, initialization probability of $P=0.5$, and $\Delta\mathbf{s}_{T}=100$.}
\label{fig:tdatasets_resenet50}
\end{figure*}

\begin{table*}[!ht]
\captionsetup{font=footnotesize}

\caption{Classification performance on test datasets. $R$ is kept trainable parameters and $\#p$ is approximate number of trainable parameters. All the values except loss and $\#p$ are in percentage. (F) refers to full network used for inference and (P) refers to pruned network using \textit{EPruning}.}

\begin{subtable}{0.49\linewidth}
\centering
\captionsetup{font=footnotesize}
\caption{ \textbf{CIFAR-10} }
\begin{adjustbox}{width=1\textwidth}
\begin{tabular}{lcccccc}
\hline
\multicolumn{1}{c}{Model}  & Loss   & Top-1   & Top-3   & Top-5   & $R$ & $\#p$      \\ \hline \hline

\multicolumn{1}{l}{ResNet-18}                    
& \multicolumn{1}{l}{0.3181} 
& \multicolumn{1}{l}{92.81} 
& \multicolumn{1}{l}{98.78} 
& \multicolumn{1}{l}{99.49}
& \multicolumn{1}{l}{100}   
& \multicolumn{1}{l}{11.2M}\\

\multicolumn{1}{l}{ResNet-18+DeepCompression}         
& \multicolumn{1}{l}{0.6951} 
& \multicolumn{1}{l}{76.15}                           
& \multicolumn{1}{l}{94.16} 
& \multicolumn{1}{l}{98.59} 
& \multicolumn{1}{l}{49.66} 
& \multicolumn{1}{l}{5.5M}   \\

\multicolumn{1}{l}{ResNet-18+EPruning(F)}        
& \multicolumn{1}{l}{0.4906} 
& \multicolumn{1}{l}{90.96} 
& \multicolumn{1}{l}{98.33} 
& \multicolumn{1}{l}{99.60} 
& \multicolumn{1}{l}{100}  
& \multicolumn{1}{l}{11.2M}\\

\multicolumn{1}{l}{ResNet-18+EPruning(P)}        
& \multicolumn{1}{l}{0.4745} 
& \multicolumn{1}{l}{90.96} 
& \multicolumn{1}{l}{98.40} 
& \multicolumn{1}{l}{99.58} 
& \multicolumn{1}{l}{49.66} 
& \multicolumn{1}{l}{5.5M}\\ \hline\hline

\multicolumn{1}{l}{ResNet-34}                    
& \multicolumn{1}{l}{0.3684} 
& \multicolumn{1}{l}{92.80} 
& \multicolumn{1}{l}{98.85} 
& \multicolumn{1}{l}{99.71} 
& \multicolumn{1}{l}{100}  
& \multicolumn{1}{l}{21.3M}  \\

\multicolumn{1}{l}{ResNet-34+DeepCompression}         
& \multicolumn{1}{l}{1.057} 
& \multicolumn{1}{l}{66.51}                           
& \multicolumn{1}{l}{91.40} 
& \multicolumn{1}{l}{97.68} 
& \multicolumn{1}{l}{38.83}   
& \multicolumn{1}{l}{8.3M}\\

\multicolumn{1}{l}{ResNet-34+EDropou(F)}         
& \multicolumn{1}{l}{0.4576} 
& \multicolumn{1}{l}{88.28} 
& \multicolumn{1}{l}{97.47} 
& \multicolumn{1}{l}{99.31} 
& \multicolumn{1}{l}{100} 
& \multicolumn{1}{l}{21.3M}  \\

\multicolumn{1}{l}{ResNet-34+EPruning(P)}        
& \multicolumn{1}{l}{0.4598} 
& \multicolumn{1}{l}{88.21}   
& \multicolumn{1}{l}{97.48}   
& \multicolumn{1}{l}{99.28}   
& \multicolumn{1}{l}{38.83} 
& \multicolumn{1}{l}{8.3M} \\ \hline\hline

\multicolumn{1}{l}{ResNet-50}                    
& \multicolumn{1}{l}{0.3761} 
& \multicolumn{1}{l}{92.21} 
& \multicolumn{1}{l}{98.70} 
& \multicolumn{1}{l}{99.51} 
& \multicolumn{1}{l}{100}   
& \multicolumn{1}{l}{23.5M} \\

\multicolumn{1}{l}{ResNet-50+DeepCompression}         
& \multicolumn{1}{l}{1.0271} 
& \multicolumn{1}{l}{67.53}                           
& \multicolumn{1}{l}{89.92} 
& \multicolumn{1}{l}{96.30} 
& \multicolumn{1}{l}{46.39}   
& \multicolumn{1}{l}{10.9M} \\ 

\multicolumn{1}{l}{ResNet-50+EPruning(F)}       
& \multicolumn{1}{l}{0.6041} 
& \multicolumn{1}{l}{85.22} 
& \multicolumn{1}{l}{96.35} 
& \multicolumn{1}{l}{98.77} 
& \multicolumn{1}{l}{100} 
& \multicolumn{1}{l}{23.5M}   \\

\multicolumn{1}{l}{ResNet-50+EPruning(P)}        
& \multicolumn{1}{l}{0.5953} 
& \multicolumn{1}{l}{85.30}   
& \multicolumn{1}{l}{96.62} 
& \multicolumn{1}{l}{98.76} 
& \multicolumn{1}{l}{46.39} 
& \multicolumn{1}{l}{10.9M} \\ \hline\hline

\multicolumn{1}{l}{ResNet-101}                  
& \multicolumn{1}{l}{0.3680} 
& \multicolumn{1}{l}{92.66} 
& \multicolumn{1}{l}{98.69} 
& \multicolumn{1}{l}{99.65} 
& \multicolumn{1}{l}{100}  
& \multicolumn{1}{l}{42.5M}   \\

\multicolumn{1}{l}{ResNet-101+DeepCompression}         
& \multicolumn{1}{l}{1.037} 
& \multicolumn{1}{l}{66.32}                           
& \multicolumn{1}{l}{92.65} 
& \multicolumn{1}{l}{98.11} 
& \multicolumn{1}{l}{45.10}   
& \multicolumn{1}{l}{19.2M} \\ 

\multicolumn{1}{l}{ResNet-101+EPruning(F)}       
& \multicolumn{1}{l}{0.6231} 
& \multicolumn{1}{l}{86.97} 
& \multicolumn{1}{l}{97.42} 
& \multicolumn{1}{l}{99.24} 
& \multicolumn{1}{l}{100}  
& \multicolumn{1}{l}{42.5M}   \\ 

\multicolumn{1}{l}{ResNet-101+EPruning(P)}       
& \multicolumn{1}{l}{0.6339} 
& \multicolumn{1}{l}{86.57} 
& \multicolumn{1}{l}{97.37} 
& \multicolumn{1}{l}{99.20} 
& \multicolumn{1}{l}{45.10} 
& \multicolumn{1}{l}{19.2M} \\ \hline\hline

\multicolumn{1}{l}{AlexNet}       
& \multicolumn{1}{l}{0.9727} 
& \multicolumn{1}{l}{84.32} 
& \multicolumn{1}{l}{96.58} 
& \multicolumn{1}{l}{99.08} 
& \multicolumn{1}{l}{100} 
& \multicolumn{1}{l}{57.4M} \\

\multicolumn{1}{l}{AlexNet+EPruning(F)}       
& \multicolumn{1}{l}{0.7632} 
& \multicolumn{1}{l}{75.05} 
& \multicolumn{1}{l}{93.74} 
& \multicolumn{1}{l}{98.18} 
& \multicolumn{1}{l}{100} 
& \multicolumn{1}{l}{57.4M} \\

\multicolumn{1}{l}{AlexNet+EPruning(P)}       
& \multicolumn{1}{l}{0.7897} 
& \multicolumn{1}{l}{74.66} 
& \multicolumn{1}{l}{93.63} 
& \multicolumn{1}{l}{97.96} 
& \multicolumn{1}{l}{77.36} 
& \multicolumn{1}{l}{44.4M} \\

\multicolumn{1}{l}{SqueezeNet}       
& \multicolumn{1}{l}{0.5585} 
& \multicolumn{1}{l}{81.49} 
& \multicolumn{1}{l}{96.31} 
& \multicolumn{1}{l}{99.01} 
& \multicolumn{1}{l}{100} 
& \multicolumn{1}{l}{0.73M} \\

\multicolumn{1}{l}{SqueezeNet+EPruning(F)}       
& \multicolumn{1}{l}{0.6686} 
& \multicolumn{1}{l}{76.76} 
& \multicolumn{1}{l}{94.55} 
& \multicolumn{1}{l}{98.62} 
& \multicolumn{1}{l}{100} 
& \multicolumn{1}{l}{0.73M} \\

\multicolumn{1}{l}{SqueezeNet+EPruning(P)}       
& \multicolumn{1}{l}{0.6725} 
& \multicolumn{1}{l}{76.85} 
& \multicolumn{1}{l}{95.00} 
& \multicolumn{1}{l}{98.56} 
& \multicolumn{1}{l}{52.35}
& \multicolumn{1}{l}{0.38M} \\  \hline
\end{tabular}
\end{adjustbox}
\label{T:results_cifar10_EPruning}
\end{subtable}
~
\begin{subtable}{0.49\linewidth}
\centering
\captionsetup{font=footnotesize}
\caption{\textbf{CIFAR-100}}
\begin{adjustbox}{width=1\textwidth}
\begin{tabular}{lcccccc}
\hline
\multicolumn{1}{c}{Model}  & Loss   & Top-1   & Top-3   & Top-5   & $R$ & $\#p$      \\ \hline \hline

\multicolumn{1}{l}{ResNet-18}                  
& \multicolumn{1}{l}{1.3830}
& \multicolumn{1}{l}{69.03} 
& \multicolumn{1}{l}{84.44} 
& \multicolumn{1}{l}{88.90} 
& \multicolumn{1}{c}{100}   
& \multicolumn{1}{l}{11.2M}\\

\multicolumn{1}{l}{ResNet-18+DeepCompression}         
& \multicolumn{1}{l}{2.3072} 
& \multicolumn{1}{l}{40.01}                           
& \multicolumn{1}{l}{62.20} 
& \multicolumn{1}{l}{72.28} 
& \multicolumn{1}{l}{48.04}  
& \multicolumn{1}{l}{5.4M} \\ 

\multicolumn{1}{l}{ResNet-18+EPruning(F)}        
& \multicolumn{1}{l}{1.9479}
& \multicolumn{1}{l}{67.04} 
& \multicolumn{1}{l}{84.11}   
& \multicolumn{1}{l}{89.43}   
& \multicolumn{1}{c}{100}   
& \multicolumn{1}{l}{11.2M}\\

\multicolumn{1}{l}{ResNet-18+EPruning(P)}    
& \multicolumn{1}{l}{1.9541} 
& \multicolumn{1}{l}{67.06} 
& \multicolumn{1}{l}{84.14} 
& \multicolumn{1}{l}{89.27}   
& \multicolumn{1}{c}{48.04}
& \multicolumn{1}{l}{5.4M}\\  \hline\hline

\multicolumn{1}{l}{ResNet-34}                  
& \multicolumn{1}{l}{1.3931} 
& \multicolumn{1}{l}{69.96} 
& \multicolumn{1}{l}{85.65}  
& \multicolumn{1}{l}{90.10} 
& \multicolumn{1}{l}{100}  
& \multicolumn{1}{l}{21.3M}  \\

\multicolumn{1}{l}{ResNet-34+DeepCompression}         
& \multicolumn{1}{l}{2.1778} 
& \multicolumn{1}{l}{42.09}                           
& \multicolumn{1}{l}{65.01}                           
& \multicolumn{1}{l}{74.31} 
& \multicolumn{1}{l}{49.41} 
& \multicolumn{1}{l}{10.5M}   \\

\multicolumn{1}{l}{ResNet-34+EPruning(F)}     
& \multicolumn{1}{l}{1.9051} 
& \multicolumn{1}{l}{64.50}  
& \multicolumn{1}{l}{81.38}   
& \multicolumn{1}{l}{86.87}   
& \multicolumn{1}{l}{100}   
& \multicolumn{1}{l}{21.3M}  \\

\multicolumn{1}{l}{ResNet-34+EPruning(P)}       
& \multicolumn{1}{l}{1.9219} 
& \multicolumn{1}{l}{64.79}  
& \multicolumn{1}{l}{81.28}  
& \multicolumn{1}{l}{86.74}   
& \multicolumn{1}{c}{49.41}  
& \multicolumn{1}{l}{10.5M}  \\ \hline\hline

\multicolumn{1}{l}{ResNet-50}                    
& \multicolumn{1}{l}{1.3068}
& \multicolumn{1}{l}{71.22}  
& \multicolumn{1}{l}{86.47}  
& \multicolumn{1}{l}{90.74}  
& \multicolumn{1}{c}{100}  
& \multicolumn{1}{l}{23.7M}  \\

\multicolumn{1}{l}{ResNet-50+DeepCompression}         
& \multicolumn{1}{l}{2.3115} 
& \multicolumn{1}{l}{43.87}                           
& \multicolumn{1}{l}{67.02} 
& \multicolumn{1}{l}{76.26} 
& \multicolumn{1}{l}{46.01}  
& \multicolumn{1}{l}{10.9M}  \\

\multicolumn{1}{l}{ResNet-50+EPruning(F)}        
& \multicolumn{1}{l}{1.8750} 
& \multicolumn{1}{l}{61.60}   
& \multicolumn{1}{l}{79.52}   
& \multicolumn{1}{l}{85.45}   
& \multicolumn{1}{c}{100}  
& \multicolumn{1}{l}{23.7M}  \\ 

\multicolumn{1}{l}{ResNet-50+EPruning(P)}    
& \multicolumn{1}{l}{1.8768} 
& \multicolumn{1}{l}{61.91}  
& \multicolumn{1}{l}{79.99}  
& \multicolumn{1}{l}{85.87} 
& \multicolumn{1}{c}{46.01} 
& \multicolumn{1}{l}{10.9M}  \\ \hline\hline 

\multicolumn{1}{l}{ResNet-101}                   
& \multicolumn{1}{l}{1.3574} 
& \multicolumn{1}{l}{71.19}   
& \multicolumn{1}{l}{85.54}   
& \multicolumn{1}{l}{90.00}   
& \multicolumn{1}{c}{100}  
& \multicolumn{1}{c}{42.6M}   \\

\multicolumn{1}{l}{ResNet-101+DeepCompression}         
& \multicolumn{1}{l}{2.6003} 
& \multicolumn{1}{l}{37.08}                           
& \multicolumn{1}{l}{58.78} 
& \multicolumn{1}{l}{68.76} 
& \multicolumn{1}{l}{43.76}  
& \multicolumn{1}{c}{18.6M}   \\

\multicolumn{1}{l}{ResNet-101+EPruning(F)}       
& \multicolumn{1}{l}{1.9558} 
& \multicolumn{1}{l}{61.52}   
& \multicolumn{1}{l}{79.71}   
& \multicolumn{1}{l}{85.20}   
& \multicolumn{1}{c}{100} 
& \multicolumn{1}{c}{42.6M}   \\

\multicolumn{1}{l}{ResNet-101+EPruning(P)}       
& \multicolumn{1}{l}{1.9412} 
& \multicolumn{1}{l}{61.92} 
& \multicolumn{1}{l}{79.49} 
& \multicolumn{1}{l}{85.23} 
& \multicolumn{1}{c}{43.76} 
& \multicolumn{1}{c}{18.6M}   \\ \hline\hline

\multicolumn{1}{l}{AlexNet}       
& \multicolumn{1}{l}{2.8113} 
& \multicolumn{1}{l}{60.12} 
& \multicolumn{1}{l}{79.18} 
& \multicolumn{1}{l}{83.31} 
& \multicolumn{1}{l}{100} 
& \multicolumn{1}{l}{57.4M} \\

\multicolumn{1}{l}{AlexNet+EPruning(F)}       
& \multicolumn{1}{l}{2.4731} 
& \multicolumn{1}{l}{56.62} 
& \multicolumn{1}{l}{78.72} 
& \multicolumn{1}{l}{81.92} 
& \multicolumn{1}{l}{100} 
& \multicolumn{1}{l}{57.4M} \\

\multicolumn{1}{l}{AlexNet+EPruning(P)}       
& \multicolumn{1}{l}{2.4819} 
& \multicolumn{1}{l}{56.59} 
& \multicolumn{1}{l}{78.52} 
& \multicolumn{1}{l}{81.62} 
& \multicolumn{1}{l}{71.84} 
& \multicolumn{1}{l}{41.2M} \\

\multicolumn{1}{l}{SqueezeNet}       
& \multicolumn{1}{l}{1.4150} 
& \multicolumn{1}{l}{67.85} 
& \multicolumn{1}{l}{85.81} 
& \multicolumn{1}{l}{89.69} 
& \multicolumn{1}{l}{100}
& \multicolumn{1}{l}{0.77M} \\

\multicolumn{1}{l}{SqueezeNet+EPruning(F)}       
& \multicolumn{1}{l}{1.5265} 
& \multicolumn{1}{l}{64.23} 
& \multicolumn{1}{l}{82.71} 
& \multicolumn{1}{l}{88.63} 
& \multicolumn{1}{l}{100}
& \multicolumn{1}{l}{0.77M} \\

\multicolumn{1}{l}{SqueezeNet+EPruning(P)}       
& \multicolumn{1}{l}{1.5341} 
& \multicolumn{1}{l}{64.02} 
& \multicolumn{1}{l}{81.63} 
& \multicolumn{1}{l}{88.51} 
& \multicolumn{1}{l}{56.40} 
& \multicolumn{1}{l}{0.43M} \\ \hline

\end{tabular}
\end{adjustbox}
\label{T:results_cifar100_EPruning}
\end{subtable}

\label{T:results_EPruning}
 \vspace{-4mm}
\end{table*}

 \section{Experiments}
 The experiments were conducted on the CIFAR-10 and CIFAR-100~\cite{krizhevsky2009learning} datasets. The horizontal flip and Cutout~\cite{devries2017improved} augmentation methods were used. Input images were resized to $32\times32$ for ResNets and $224\times224$ for AlexNet~\cite{krizhevsky2012imagenet} and SqueezeNet v1.1~\cite{iandola2016squeezenet}. 
 
 We used ResNets (18, 34, 50, and 101 layers)~\cite{he2016deep}, AlexNet~\cite{krizhevsky2012imagenet}, SqueezeNet v1.1~\cite{iandola2016squeezenet}, and Deep Compression~\cite{han2015deep} to evaluate EPruning. The results were averaged over five independent runs. A grid hyper-parameter search was conducted based on the Top-1 accuracy for all models, including initial learning rates in $\{0.01,0.1,1\}$, Stochastic Gradient Descent (SGD)~\cite{hinton2012neural} and Adadelta~\cite{zeiler2012adadelta} optimizer, exponential and step learning rate decays with gamma values in $\{25,50\}$, and batch sizes of 64 and 128. The Adadelta optimizer with Step adaptive learning rate (step: every 50 epoch at gamma rate of 0.1) and weight decay of $10e^{-6}$ was used. The number of epochs was 200 and the batch size was 128. Random dropout was not used in the \textit{EPruning} experiments. For the other models, where applicable, the random dropout rate was set to 0.5. The early state convergence in~(\ref{eq:stateconvergence}) is used with a threshold of 100 epochs. The models are implemented in PyTorch~\cite{NEURIPS2019_9015} and trained on three NVIDIA Titan RTX GPUs.

 Table~\ref{T:results_EPruning} shows the classification performance results. The original models contain the entire trainable parameters and have larger learning capacity. \textit{EPruning} in pruned and full versions have slightly lower Top-1 performance than the original model and competitive performance in terms of Top-5 performance. The Deep Compression~\cite{han2015deep} method receives pruning rate as input. For the sake of comparison, we have modified it to perform pruning on every convolution layer, given the rate achieved by \textit{EPruning}, where generally it has lower performance than \textit{EPruning}. SqueezeNet~\cite{SqueezeNet} is a small network with AlexNet level accuracy. \textit{EPruning} is also applied to AlexNet and SqueezeNet v1.1, where it has a smaller pruning rate for AlexNet compared to ResNets but can prune approximately half of the trainable parameters in SqueezeNet v1.1 and achieve slightly lower performance.

Figure~\ref{fig:tdatasets_resenet50} shows convergence plots of ResNet-50 with respect to the cross-entropy loss and energy of the sub-network over 200 training epochs. The plots show that the best energy is lower than the average energy and after epoch 100, due to early stopping, the sub-network is selected. Training loss and energy loss follow similar declining trend and after epoch 100, the training loss has a slower declining rate. We observe that before epoch 100 the model was in an exploration phase and after this epoch enters an exploitation (fine-tuning) phase.

This method increases the computational complexity of the model due to evaluation of energy loss for each candidate state vector in the population. However, with proper parallel implementation at each state vector level and shared memory management, this overhead can significantly be reduced. As an example, for ResNet-50, the inference time per image is 5.50E-5 Seconds where with EPruning with a population size of 8 this time is 3.40E-4 Seconds and in parallel mode this time is 8.95E-5 Seconds.

\section{Conclusions}
\label{sec:conclusion}
In this paper, we have proposed a stochastic framework for pruning deep neural networks. 
Most pruning and compression models first prune a network and then fine-tune it. The proposed \textit{EPruning} method has two phases. The first phase acts as dropout where various subsets of the neural network are trained. Each sub-network is selected based on a corresponding energy loss value which reflects performance of the sub-network. The second phase is a fine-tuning phase with focus on training the pruned network. The proposed framework has the advantage of immediate usability for any neural network without manual modification of layers. In addition, predefined number of active states can also be utilized in the optimizer to enforce a specific dropout/pruning rate. Our experiments show that as the proposed framework searches for sub-networks with lower energy, the training loss also decreases.

\section{Acknowledgment}
The authors acknowledge financial support of Fujitsu Laboratories Ltd. and Fujitsu Consulting (Canada) Inc. 
\bibliographystyle{IEEEbib}
\bibliography{strings,mybibfile}

\end{document}